\def\year{2021}\relax
\documentclass[letterpaper]{article} 
\usepackage{aaai21}  
\usepackage{times}  
\usepackage{helvet} 
\usepackage{courier}  
\usepackage[hyphens]{url}  
\usepackage{graphicx} 
\usepackage{natbib}  
\usepackage{caption} 
\usepackage{amssymb}
\usepackage{amsthm}
\usepackage{amsfonts} 
\usepackage{subfigure}
\usepackage{graphicx}
\usepackage{amsthm}
\usepackage{mathtools}
\usepackage{cuted}
\usepackage{makecell}
\usepackage{multirow}
\usepackage{booktabs}
\usepackage{color}
\newtheorem{theorem}{Theorem}

\title{Fractal Autoencoders for Feature Selection}
\author {
    Xinxing Wu,
    Qiang Cheng\thanks{Corresponding author}\\
}
\affiliations {
    University of Kentucky, Lexington, Kentucky, USA\\
    xinxingwu@gmail.com, qiang.cheng@uky.edu
}
\begin{document}

\maketitle

\begin{abstract}
Feature selection reduces the dimensionality of data by identifying a subset of the most informative features. In this paper, we propose an innovative framework for unsupervised feature selection, called fractal autoencoders (FAE). It trains a neural network to pinpoint informative features for global exploring of representability and for local excavating of diversity. Architecturally, FAE extends autoencoders by adding a one-to-one scoring layer and a small sub-neural network for feature selection in an unsupervised fashion. With such a concise architecture, FAE achieves state-of-the-art performances; extensive experimental results on fourteen datasets, including very high-dimensional data, have demonstrated the superiority of FAE over existing contemporary methods for unsupervised feature selection. In particular, FAE exhibits substantial advantages on gene expression data exploration, reducing measurement cost by about $15$\% over the widely used L1000 landmark genes. Further, we show that the FAE framework is easily extensible with an application.
\end{abstract}

\section{Introduction}
\noindent High-dimensional data is pervasive in almost every area of modern data science~\citep{Clarke,Blum}. Dealing with high-dimensional data is challenging due to the known phenomenon -- the curse of dimensionality~\citep{Bellman}. In numerous applications, principal component analysis (PCA)~\citep{Pearson} and autoencoders (AE)~\citep{David,Ballard}, two traditional and simple approaches, are typically used for dimensionality reduction. For example, PCA is adopted to reduce the dimensions of gene expression data before the extraction of the samples' rhythmic structures~\citep{Ron}; AE are employed to process high-dimensional datasets prior to clustering~\citep{Xie} and subspace learning~\citep{Ji}. Despite their widespread usage, the interpretation of lower-dimensional feature spaces produced by PCA and AE is not straightforward, because these feature spaces are different from the original feature space. In contrast to PCA and AE, feature selection allows for ready interpretability with the input features, by identifying and retaining a subset of important features directly from the original feature space~\citep{Guyon}.

There exist various feature selection approaches. According to whether labels are used, they can be categorized as supervised~\citep{cheng2010fisher}, semi-supervised, and unsupervised methods~\citep{Alelyani}. Unsupervised approaches have potentially extensive applications, since they do not require labels that can be rare or expensive to obtain. A variety of techniques for unsupervised feature selection have been proposed, e.g., Laplacian score (LS)~\citep{He} and concrete autoencoders (CAE)~\citep{Abubakar}. While often used, the existing approaches may still exhibit suboptimal performance in downstream learning tasks on many datasets, which can be seen, e.g., in Table~\ref{table3}. There are two major reasons that cause such under-performance. First, the space to search for potentially important subsets of features in the absence of the guidance by labels is often very large, which renders unsupervised feature selection to be like finding a needle in a haystack. Second, it is necessary, yet challenging, to take account of the inter-feature interactions. Ideally, the selected features should be globally representative and as diverse as possible. If the selected features are all important yet highly correlated, they may be capable of representing only partial data, and thus they would hardly comprise a globally representative feature subset. For example, if a pixel of a natural image is important, then some neighboring ones are also likely to be so because of the typical spatial dependence in images; thus, to select diverse, salient features to represent the overall contents, if a pixel is important and selected, those neighboring pixels of high correlations with it should not be included into the feature subset. Existing unsupervised approaches have limited abilities to simultaneously explore the large search space for features that can represent the overall contents and take into account the diversity, which is reflected by the inter-correlation of features, thus leading to suboptimal performances.

To overcome these difficulties, in this paper we propose a novel unsupervised feature selection framework, called fractal autoencoders (FAE). It trains a neural network (NN) to identify potentially informative features for representing the contents globally; simultaneously, it exploits a dependence sub-NN to select a subset locally from the globally informative features to examine their diversity, which is efficiently measured by their abilities to reconstruct the original data. In this way, the sub-NN enables FAE to effectively screen out the highly correlated features; the global AE component of FAE turns out to play a crucial role of regularization to stabilize the feature selecting process, aside from its standard role of feature extraction. With our new architecture, FAE merges feature selection and feature extraction into one model, facilitating the identification of a subset of the most representative and diverse input features. To illustrate the extensive ability of the FAE framework, we use it to derive an $h$-Hierarchy FAE ($h$-HFAE) application to identity multiple subsets of important features.

In summary, our main contributions include: 
\begin{itemize}
\item We propose a novel framework, FAE, for feature selection to meet the challenges of existing methods. It combines global exploration of representative features and local excavation of their diversity, thereby enhancing the generalization of the selected feature subset and accounting for inter-feature correlations.
\item As our framework can be readily applicable or extensible to other tasks, we show an application to identity multiple hierarchical subsets of salient features simultaneously. 
\item We validate FAE with extensive experiments on fourteen real datasets. Although simple, it demonstrates state-of-the-art performance for reconstruction on many benchmarking datasets. It also yields superior performance in a downstream learning task of classification on most of the benchmarking datasets. As a biological application, FAE reduces gene expression measurements by about $15$\% compared with L1000 landmark genes. Further, FAE exhibits more stable performance on varying numbers of selected features than contemporary methods.
\end{itemize}

The notations and definitions are given as follows. Let $n$, $m$, $k$, and $d$ be the numbers of samples, features, selected features, and reduced dimensions, respectively. Let $\mathbf{X}\in\mathbb{R}^{n\times m}$ be a matrix containing the input data. A bold capital letter such as $\mathbf{W}$ denotes a matrix; a lowercase bold capital letter such as $\mathbf{w}$ denotes a vector; Diag($\mathbf{w}$) represents a diagonal matrix with the diagonal $\mathbf{w}$; $\mathbf{w}^{\mathrm{max}_k}$ is an operation to keep the $k$ largest entries of $\mathbf{w}$ while making other entries $0$. $\|\cdot\|_{\mathrm{F}}$ denotes the Frobenius norm. 

The remaining of the paper is organized as follows. We first discuss the related work, then present our proposed approach, followed by extensive experiments. Finally, we apply FAE to identify multiple subsets of informative features.

\section{Related Work}
A variety of feature selection approaches have been proposed. They are usually classified into four categories~\citep{Alelyani,Li}: filter methods, which are independent of learning models; wrapper methods, which rely on learning models for selection criteria; embedder approaches, which embed the feature selection into learning models to also achieve model fitting simultaneously; hybrid approaches, which are a combination of more than one of above three. Alternatively, the approaches are categorized as supervised, semi-supervised, and unsupervised methods according to whether label information is utilized. Unsupervised feature selection has potentially broad applications because it requires no label information~\citep{peng2016feature,peng2017nonnegative}; yet, it is also arguably more challenging due to the lack of labels to guide the identification of relevant features. In this paper, we focus on unsupervised feature selection and briefly review typical methods below.

{LS~\citep{He} is a filter method that uses the nearest neighbor graph to model the local geometric structures of the data. By selecting the features which are locally the smoothest on the graph, LS focuses on the local property yet neglects the global structure. SPEC~\citep{Zhao} is a filter method based on general similarity matrix. It employs the spectrum of the graph to measure feature relevance and unifies supervised and unsupervised feature selection. Principal feature analysis (PFA)~\citep{Lu} utilizes the structure of the principal components of a set of features to select the subset of relevant features. It can be regarded as a wrapper method to optimize the PC coefficients, and it mainly focuses on globality. Multi-cluster feature selection (MCFS)~\citep{Cai} selects a subset of features to cover the multi-cluster structure of the data, where spectral analysis is used to find the inter-relationship between different features. Unsupervised discriminative feature selection (UDFS)~\citep{Yang} incorporates the discriminative analysis and $\ell_{2,1}$ regularization to identify the most useful features. Nonnegative discriminative feature selection (NDFS)~\citep{Zechao} jointly learns the cluster labels and feature selection matrix to select discriminative features. It uses a nonnegative constraint on the class indicator to learn cluster labels and adopts an $\ell_{2,1}$ constraint to reduce the redundant or noisy features. Infinite feature selection (Inf-FS)~\citep{Roffo} implements feature selection by taking into account all the possible feature subsets as paths on a graph, and it is also a filter method.

Recently, a few AE-based feature selection methods have been developed. Autoencoder feature selector (AEFS)~\citep{Han} combines autoencoders regression and $\ell_{2,1}$ regularization on the weights of the encoder to obtain a subset of useful features. It exploits both linear and nonlinear information in the features. Agnostic feature selection (AgnoS)~\citep{Doquet} adopts AE with explicit objective function regularizations, such as the $\ell_{2,1}$ norm on the weights of the first layer of AE (AgnoS-W), $\ell_{2,1}$ norm on the gradient of the encoder (AgnoS-G), and $\ell_1$ norm on the slack variables that constitute the first layer of AE (AgnoS-S), to implement feature selection. AgnoS-S is the best of the three, so in this study we will compare our approach with AgnoS-S. CAE~\citep{Abubakar} replaces the first hidden layer of AE with a \lq\lq concrete selector\rq\rq\,\,layer, which is the relaxation of a discrete distribution called concrete distribution~\citep{Maddison}, and then it picks the features with an extremely high probability of connecting to the nodes of the concrete selection layer. The parameters of this layer are estimated by the reparametrization trick~\citep{Kingma}. CAE reports superior performance over other competing methods. 


MCFS, UDFS, NDFS, AEFS, AgnoS, and CAE can be all regarded as embedded approaches. Though our proposed FAE model also embeds the feature selection into AE, which looks similar to AgnoS, AEFS, and CAE, it essentially differs from these existing methods: AEFS and AgnoS mainly depend on exploiting sparsity norm regularizations such as $\ell_{2,1}$ and $\ell_{1}$ on the weights of AE to select features, which do not consider diversity; in contrast, FAE consists of two NNs, and it innovatively adopts a sub-NN to explicitly impose the desired diversity requirement on informative features. CAE adopts a probability distribution on the first layer of AE and selects features by their parameters. However, several neurons in the concrete selector layer may potentially select the same or redundant features, and the training requires that the average of the maximum probability of connecting to these neurons in the concrete selection layer exceed a pre-specified threshold close to $1$, which may be hard to attain for high-dimensional datasets; meanwhile, the second and third top features at different nodes of the concrete selector layer may be insignificant because of their trivial average probability. These potential drawbacks can limit the performance of CAE. In contrast, the proposed FAE does not depend on any probability distribution; rather, its sub-NN, with the guidance by the global-NN, directly pinpoints a subset of selected features, which makes FAE concise in architecture and easily applicable to different tasks.  
\section{Proposed Approach}
In this section, we will present the architecture and formulation of FAE.

\subsection{Overview of Our Approach}
The architecture of our FAE approach is depicted in Figure~\ref{fig:02}. It enlists the AE architecture as a basic building block; yet, its structure is particularly tailored to feature selection. In the  following, we will explain the architecture and its components in detail.  
\begin{figure}[t] 
  \centering
	\includegraphics[width=0.46\textwidth]{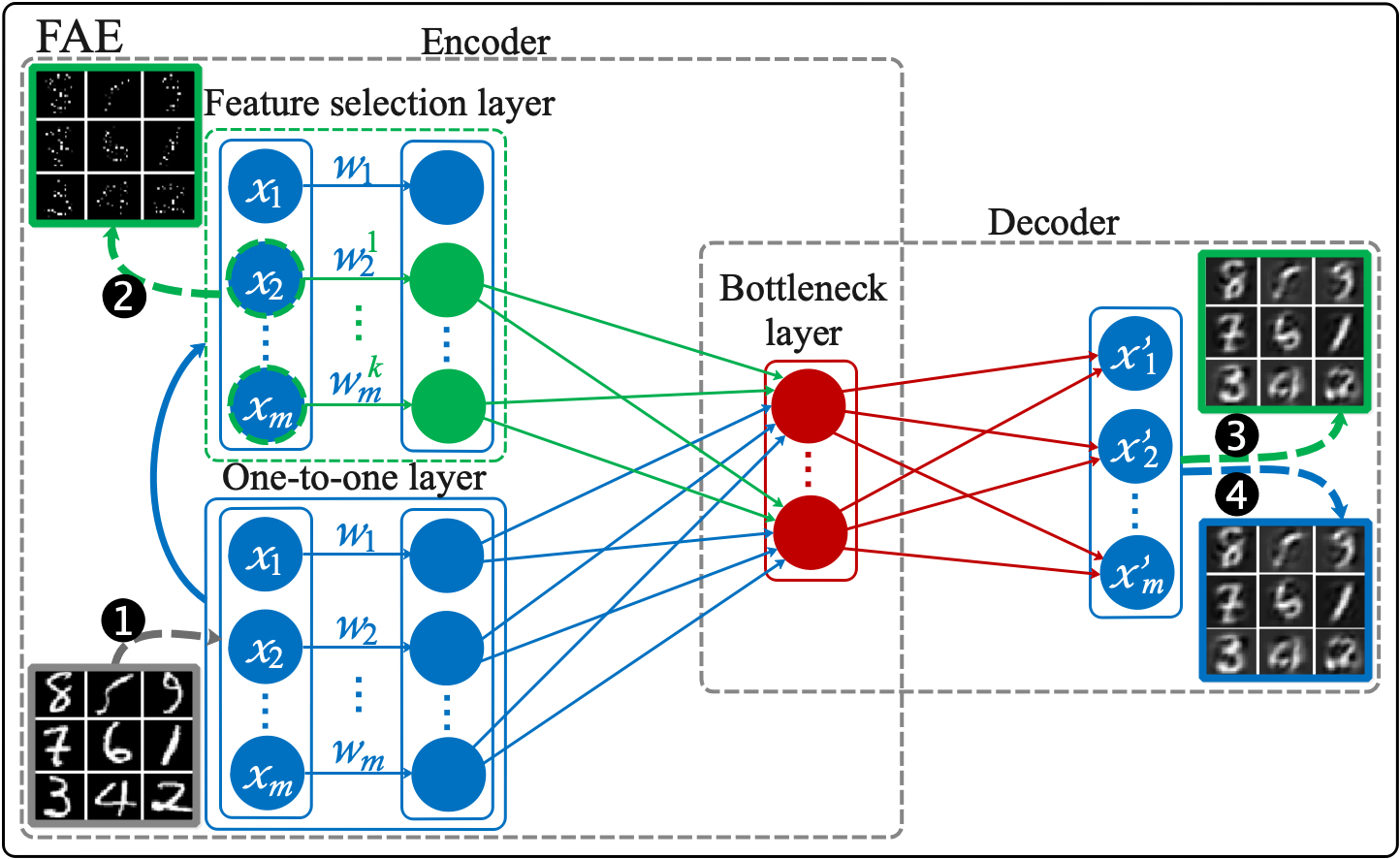}
  \caption{The architecture of FAE. During training, the global NN (with one-to-one layer) and its dependence sub-NN (with feature selection layer) are used to optimize (\ref{FAE}); during testing, only the trained sub-NN is used to select features and reconstruct the data. Potentially, FAE implements feature extraction. The presented quantifies are: 1) input; 2) feature selection result; 3) reconstruction based on the selected features, i.e., $f(g(\mathbf{X}\mathbf{W}_{\mathrm{I}}^{\mathrm{max}_k}))$; 4) reconstruction from the one-to-one layer, i.e., $f(g(\mathbf{X}\mathbf{W}_{\mathrm{I}}))$.}
\label{fig:02}
\end{figure}  

\subsection{Formalization of Autoencoders}

For AE, we formalize it as follows:
\begin{equation}{\label{AE}}
\displaystyle\min_{f,g}\|\mathbf{X}-f(g(\mathbf{X}))\|_{\mathrm{F}}^2,
\end{equation}
where $g$ is an encoder, and $f$ is a decoder. $g(\mathbf{X})$ embeds the input data into a latent space $\mathbb{R}^{n\times d}$, where $d$ is the dimension of the bottleneck layer of AE. Taking MNIST as an example, for $d=49$, we visualize the encoded samples in Figure~\ref{fig:01} (b). After being transformed, either nonlinearly or linearly, from the original space, the contents of each sample are not visually meaningful in the latent space. 

\subsection{Formalization of Unsupervised Feature Selection}

Feature selection is to identify a subset of informative features in the original feature space, and it can be formalized as follows:
\begin{equation}{\label{FS}}
\displaystyle\min_{S^k,H}\|H(\mathbf{X}_{S^k})-\mathbf{X}\|_{\mathrm{F}}^2,
\end{equation}
where $S^k$ denotes the subset of the specified $k$ features, $\mathbf{X}_{S^k}$ is the derived data set from $\mathbf{X}$ based on $S^k$, and $H$ denotes a mapping from the space spanned by $\mathbf{X}_{S^k}$ to ${\mathbb{R}}^{n \times m}$ in the absence of information about labels. The optimization problem in (\ref{FS}) is NP-hard~\citep{Natarajan,Hamo}. This paper will develop an effective algorithm to approximate the solution of (\ref{FS}) for unsupervised feature selection.

\subsection{Identification Autoencoders (IAE)}
To perform feature selection in the original space, our first attempt is to add a simple one-to-one layer between the input and hidden layers of AE to weigh the importance of each input feature. It is also natural to exploit the sparsity property of $l_1$ regularization for the weights of this layer for feature selection, inspired by Lasso~\citep{Tibshirani}. Then, we have the following formulation:
\[
\displaystyle\min_{\mathbf{W}_{\mathrm{I}},f,g}\|\mathbf{X}-f(g(\mathbf{X}\mathbf{W}_{\mathrm{I}}))\|_{\mathrm{F}}^2+\lambda_1\|\mathbf{W}_{\mathrm{I}}\|_{1},\,\, \mathrm{s.t.}\,\,\mathbf{W}_{\mathrm{I}}\geqslant 0,
\]
where $\mathbf{W}_{\mathrm{I}}=$ Diag($\mathbf{w}$), $\mathbf{w}\in{\mathbb{R}}^m$, and $\lambda_1$ is a parameter balancing between the reconstruction error and sparsity regularization. The $\ell_{1}$ norm induces sparsity and shrinks the less important features’ weights to $0$, and it may make the features more discriminative as well. Here, we require that the entries of $\mathbf{W}_{\mathrm{I}}$ should be nonnegative since they represent the importance of the features and the non-negativity constraint would make their interpretation more meaningful~\citep{Xu}. The fully connected concrete layer~\citep{Abubakar} has taken a similar non-negativity constraint, albeit for a full matrix.   

For AE with such an additional one-to-one layer and the modified objective function, we call it an identification autoencoder (IAE) only for notational purpose. Actually, IAE is a general case of AgnoS-S~\citep{Doquet}, where it does not impose any constraint on the dimension of the bottleneck layer of AE. After training, the features corresponding to the $k$ largest entries of $\mathbf{W}_{\mathrm{I}}$ are selected as the most informative features. Compared with standard AE, IAE clearly increases no more than $m$ additional parameters.

We may visualize the selected features by IAE in Figure~\ref{fig:01} (c) and (e). It is seen that IAE captures a part of key features from the original samples; however, it cannot capture other key features on the skeleton of the digits, and the selected features fail to recover the original contents, as shown in Figure~\ref{fig:01} (g). In general, the selected features by unsupervised feature selection are to be representative of the input data, implying that the selected features should reconstruct the original samples well. Thus, IAE cannot serve the purpose of feature selection in itself. Its failure is mainly due to the lack of diversity of its  selected features. The $\ell_1$ regularization term in IAE may promote the sparsity of the feature weight vector; however, it cannot ensure a sufficient level of diversity needed by a representative subset of features. Because the features in real data often have significant inter-correlations and even redundancy, without properly taking account of them, the selected features would have high correlations yet lack necessary diversity. Directly computing the pairwise interactions of all features requires a full $m \times m$ weight matrix, which may be computationally costly for high-dimensional data. Accounting for higher-order interactions between features would require even higher complexities. To address this problem, we propose a simple yet effective approach by using a sub-NN to locally excavate for diversity information from the feature weights, thereby reducing the search space significantly. We will introduce this sub-network below, which leads to the architecture of FAE.  

\subsection{Fractal Autoencoders (FAE)}
To remedy the diversity issue of IAE, we further design a sub-NN term, which requires that the subset of $k$ selected features from $\mathbf{W}_{\mathrm{I}}$ should be so diverse as to still represent the global contents of original samples as much as possible. Putting together, our proposed model is as follows:
\begin{equation}{\label{FAE}}
\begin{array}{l}
\displaystyle\min_{\mathbf{W}_{\mathrm{I}},f,g}\|\mathbf{X}-f(g(\mathbf{X}\mathbf{W}_{\mathrm{I}}))\|_{\mathrm{F}}^2+\lambda_1\|\mathbf{X}-f(g(\mathbf{X}\mathbf{W}_{\mathrm{I}}^{\mathrm{max}_k}))\|_{\mathrm{F}}^2\\
\displaystyle+\lambda_2\|\mathbf{W}_{\mathrm{I}}\|_{1},\,\, \mathrm{s.t.}\,\,\mathbf{W}_{\mathrm{I}}\geqslant 0,
\end{array}
\end{equation}
where $\mathbf{W}_{\mathrm{I}}^{\mathrm{max}_k}=$ Diag$(\mathbf{w}^{\mathrm{max}_k})$, and $\lambda_1$ and $\lambda_2$ are nonnegative balancing parameters. We call the neural network corresponding to (\ref{FAE}) fractal autoencoders (FAE), due to its seemingly self-similarity characteristic: A small proportion of features in the second term achieve a similar performance to the whole set of features in the first term for reconstructing the original data. This characteristic will be manifested more clearly when applying FAE to extract multiple subsets of features later. 

In training, we solve $\mathbf{W}_{\mathrm{I}}^{\mathrm{max}_k}$ by jointly optimizing the global-NN and sub-NN. After training FAE, we obtain $\mathbf{W}_{\mathrm{I}}^{\mathrm{max}_k}$ which can be used to perform feature selection on new samples during testing. We illustrate the selected features, the selected features superimposed on the original samples (for easy visualization), and reconstructed samples with these features in (d), (f), and (h) of Figure~\ref{fig:01}, respectively, for 9 random samples from MNIST. 
\begin{figure}[t]
\centering 
\subfigure[Original testing samples]{
\centering
{
\begin{minipage}[t]{0.4\linewidth}
\centering
\centerline{\includegraphics[width=1.2\textwidth]{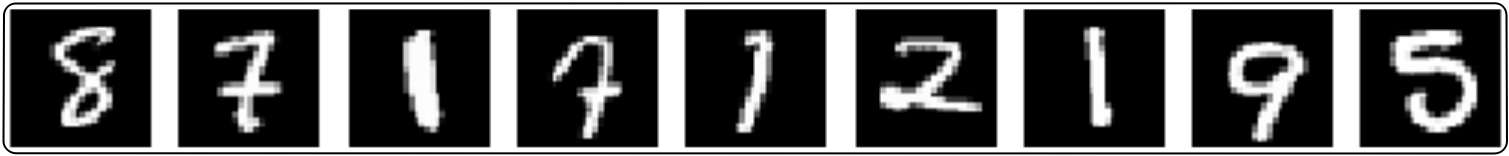}}
\end{minipage}%
}%
}%
\hspace{0.24in}
\subfigure[Features from AE]{
\centering
{
\begin{minipage}[t]{0.4\linewidth}
\centering
\centerline{\includegraphics[width=1.2\textwidth]{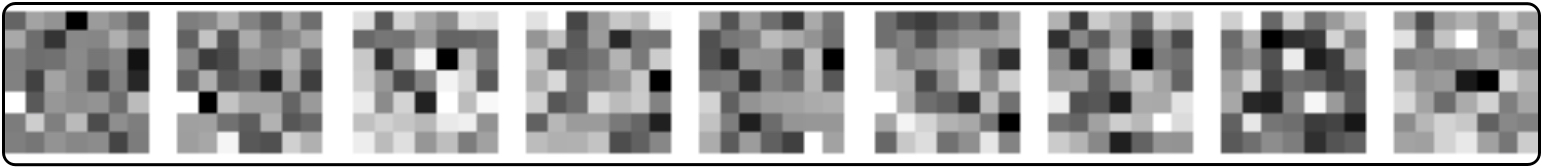}}
\end{minipage}%
}%

}%
\vskip -0.01in
\subfigure[Features from IAE]{
\centering
{
\begin{minipage}[t]{0.4\linewidth}
\centering
\centerline{\includegraphics[width=1.2\textwidth]{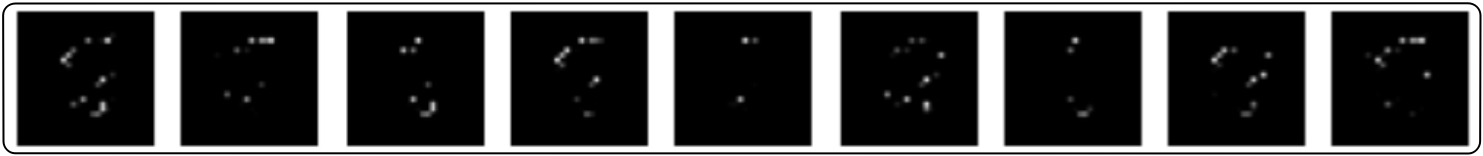}}
\end{minipage}%
}%
}%
\hspace{0.24in}
\subfigure[Features from FAE]{
\centering
{
\begin{minipage}[t]{0.4\linewidth}
\centering
\centerline{\includegraphics[width=1.2\textwidth]{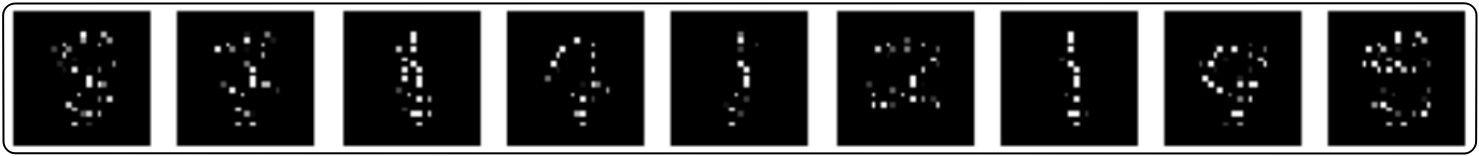}}
\end{minipage}%
}%

}%
\vskip -0.01in
\subfigure[Key features by IAE]{
\centering
{
\begin{minipage}[t]{0.4\linewidth}
\centering
\centerline{\includegraphics[width=1.2\textwidth]{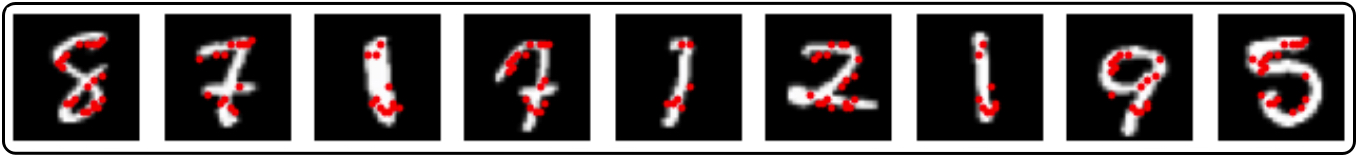}}
\end{minipage}%
}%
}%
\hspace{0.24in}
\subfigure[Key features by FAE]{
\centering
{
\begin{minipage}[t]{0.4\linewidth}
\centering
\centerline{\includegraphics[width=1.2\textwidth]{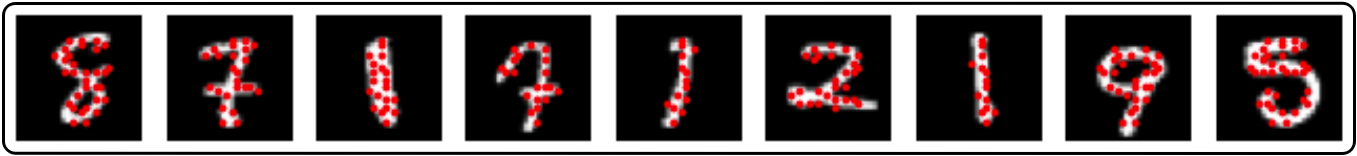}}
\end{minipage}%
}%

}%
\vskip -0.01in
\subfigure[Reconstruction by IAE]{
\centering
{
\begin{minipage}[t]{0.4\linewidth}
\centering
\centerline{\includegraphics[width=1.2\textwidth]{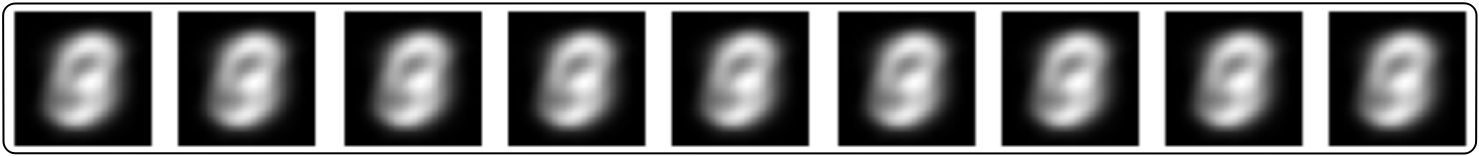}}
\end{minipage}%
}%
}%
\hspace{0.24in}
\subfigure[Reconstruction by FAE]{
\centering
{
\begin{minipage}[t]{0.4\linewidth}
\centering
\centerline{\includegraphics[width=1.2\textwidth]{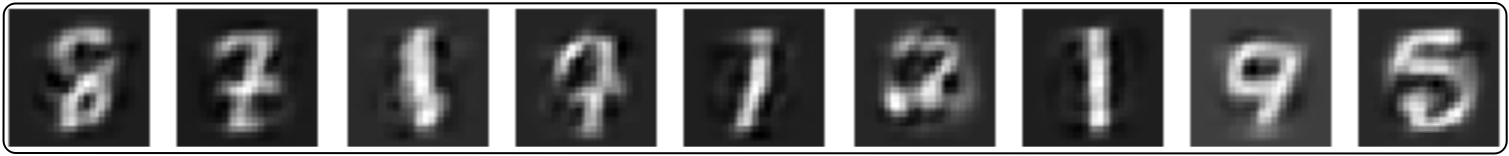}}
\end{minipage}%
}%

}%
\centering
\caption{(a) Testing samples randomly chosen from MNIST; (b) 49 features extracted by AE (size enlarged for visualization); (c) 50 features selected by IAE; (d) 50 features selected by FAE; (e) key features by IAE shown with original samples; (f) key features by FAE shown with original samples; (g) IAE's reconstruction based on the 50 features; (h) FAE's reconstruction based on the 50 features; (c)-(h) are best viewed with enlarging.}
\label{fig:01}
\end{figure}

\section{Experiments}
In this section, we will perform experiments to extensively assess FAE by comparing it with contemporary methods on many benchmarking datasets. 
\begin{table}[!htbp]
\centering
\resizebox{0.85\columnwidth}{!}{
  {\setlength{\aboverulesep}{0pt}
   \setlength{\belowrulesep}{0pt}
  \begin{tabular}{ll!{\vrule width0.8pt}lll}
    \toprule
    &Dataset&\# Sample&\# Feature/\# Gene&\# Class\\
    \midrule
    1&Mice Protein		& 1,080 		& 77 			& 8\\ 
    2&COIL-20		 	& 1,440 		& 400 			& 20\\
    3&Activity		 	& 5,744 		& 561 			& 6\\ 
    4&ISOLET		 	& 7,797 		& 617 			& 26\\  
    5&MNIST		 		& 10,000 	& 784 			& 10\\ 
    6&MNIST-Fashion		& 10,000 	& 784 			& 10\\
    7&USPS				& 9,298		& 256			& 10\\
    8&GLIOMA 				& 50		& 4,434			& 4\\
    9&leukemia			& 72		& 7,070 			& 2\\
   10&pixraw10P			& 100		& 10,000 		& 10\\
   11&Prostate$\_$GE		& 102		& 5,966			& 2\\
   12&warpAR10P			& 130		& 2,400			& 10\\
   13&SMK$\_$CAN$\_$187   & 187		& 19,993			& 2\\
   14&arcene				& 200 		& 10,000			& 2\\
   15&GEO          		& 111,009 	& 10,463 		& Null\\
    \bottomrule
  \end{tabular}}}
  \caption{Statistics of datasets.}
\label{table1}
\end{table}
\subsection{Datasets to Be Used}
The benchmarking datasets used in this paper are Mice Protein Expression~\citep{UCI2}, COIL-20~\citep{COIL20}, Smartphone Dataset for Human Activity Recognition in Ambient Assisted Living~\citep{Anguita}, ISOLET~\citep{UCI1}, MNIST~\citep{LeCun}, MNIST-Fashion~\citep{MNISTFashion}, GEO{\footnote{Obtained from https://cbcl.ics.uci.edu/public$\_$data/D-GEX/~\citep{Chen}.}}, USPS, GLIOMA, leukemia, pixraw10P, Prostate$\_$GE, warpAR10P, SMK$\_$CAN$\_$187, and arcene{\footnote{The last eight datasets are from the scikit-feature feature selection repository~\citep{Li}.}}. We summarize the statistics of these datasets in Table~\ref{table1}. Following CAE~\citep{Abubakar} and considering the long runtime of UDFS, for MNIST and MNIST-Fashion, we randomly choose $6,000$ samples from each training set to train and validate and $4,000$ from each testing set for testing. And we randomly split $6,000$ samples into training and validation sets at a ratio of $90:10$. For GEO, we randomly split the preprocessed GEO in the same way as D-GEX~\citep{Chen}: $88,807$ for training, $11,101$ for validating, and $11,101$ for testing{\footnote{\citet{Abubakar} stated that they used the same preprocessing scheme with D-GEX. Though having the same number of features, we note that their dataset has a slightly different sample size $112,171$ from ours and that in~\citep{Chen}.}}. For other datasets, we randomly split them into training, validation, and testing sets by a ratio of $72:8:20$.

\subsection{Design of Experiments}
In experiments of FAE, we set the maximum number of epochs to be $1,000$ for datasets 1-14 and $200$ for dataset 15. We initialize the weights of feature selection layer by sampling uniformly from $\mathrm{U}[0.999999, 0.9999999]$ and the other layers with the Xavier normal initializer. We adopt the Adam optimizer~\citep{Kingma1} with an initialized learning rate of $0.001$. We set $\lambda_1$ and $\lambda_2$ in \eqref{FAE} to $2$ and $0.1$, respectively. For the hyper-parameter setting, we perform a grid search on the validation set, and then choose the optimal one. In the following experiments, we only use the linear version of FAE for simplicity, that is, $g(\mathbf{X})=\mathbf{X}\mathbf{W}_{\mathrm{E}}$, $\mathbf{W}_{\mathrm{E}}\in\mathbb{R}^{m\times k}$, and $f(g(\mathbf{X}))=(g(\mathbf{X}))\mathbf{W}_{\mathrm{D}}$, $\mathbf{W}_{\mathrm{D}}\in\mathbb{R}^{k\times m}$. The simple, linear version of FAE can already achieve superior performance, as shown below. 

For the specified number of selected features $k$, we adopt two options: 1) We take $k=10$ for Mice Protein dataset, $50$ for datasets $2$-$7$ following CAE~\citep{Abubakar}, and $64$ for high-dimensional datasets $8$-$14$. For all baseline methods, we adopt this option. 2) For FAE, we additionally use fewer features, with $k=8$ for Mice Protein dataset, $36$ for datasets $2$-$7$, and $50$ for datasets $8$-$14$, to further show its superior representative ability over competing methods. We set the dimension of the latent space to $k$ and denote FAE with these two options as Opt1 and Opt2, respectively. 

Two metrics are used for evaluating the models: 1) reconstruction error, which is measured in mean squared error (MSE); 2) classification accuracy, which is measured by passing the selected features to a downstream classifier as a viable means to benchmark the quality of the selected subset of features. For fair comparison, following CAE~\citep{Abubakar}, after selecting the features, we train a linear regression model with no regularization to reconstruct the original features, and the resulting linear reconstruction error is used as the first metric{\footnote{For reconstruction error only, it denotes the error from the second term of~\eqref{FAE}, that is, $\|\mathbf{X}-f(g(\mathbf{X}\mathbf{W}_{\mathrm{I}}^{\mathrm{max}_k}))\|_{\mathrm{F}}^2$.}}. Meanwhile, for the second metric we use the extremely randomized trees~\citep{Geurts} as the classifier. 

All experiments are implemented with Python 3.7.8, Tensorflow 1.14, and Keras 2.2.5. The codes can be found at https://github.com/xinxingwu-uk/FAE.

\subsection{Results on Fourteen Datasets}
The experimental results on reconstruction and classification with the selected features by different algorithms are reported in Tables~\ref{table2} and~\ref{table3}. For all the results, we implement $5$ runs with random splits on the fixed dataset to present mean results and standard errors. From Table~\ref{table2}, it is seen that FAE yields smaller reconstruction errors than baseline methods on majority datasets, indicating its strong ability for representing the original data. From Table~\ref{table3}, it is evident that FAE exhibits consistently superior performance in the downstream classification task on most of the benchmarking datasets. 

\begin{table*}[!htbp] 
\centering
\resizebox{1.9\columnwidth}{!}{
  {\setlength{\aboverulesep}{0pt}
   \setlength{\belowrulesep}{0pt}
  \begin{tabular}{l!{\vrule width0.8pt}cccccccccc|cc}
    \toprule
    \multirow{2}*{{\Large Dataset}}&\multirow{2}*{{\Large LS}}&\multirow{2}*{{\Large SPEC}}&\multirow{2}*{{\Large NDFS}}&\multirow{2}*{{\Large AEFS}}&\multirow{2}*{{\Large UDFS}}&\multirow{2}*{{\Large MCFS}}&\multirow{2}*{{\Large PFA}}&\multirow{2}*{{\Large Inf-FS}}&\multirow{2}*{{\Large AgnoS-S}}&\multirow{2}*{{\Large CAE}}&\multicolumn{2}{c}{{\Large FAE}}\\
    					& 		& 		& 		 & 	  		 &	   &	  & 	&	&		& 	  		&Opt1		&Opt2\\
    \midrule
    Mice Protein 		& .575$\pm$.118 &  1.32$\pm$1.78 &  1.69$\pm$0.98 & .020$\pm$.007 &  {\bf .009$\pm$.006} &  16.5$\pm$30.4 &  .028$\pm$.002 & .443$\pm$.040 &  .038$\pm$.014 &  .032$\pm$.001 &  .014$\pm$.005 &  .015$\pm$.005\\
    COIL-20      		& .225$\pm$.035 &  .711$\pm$.626 &  .144$\pm$.016 &  .011$\pm$.0 &  .015$\pm$.001 &  2.89$\pm$2.47 &  {\bf .009$\pm$.0} & .134$\pm$.013 &  .035$\pm$.009 &  .011$\pm$.001 &  .011$\pm$.001 &  .013$\pm$.001 \\
    Activity     		& 4166$\pm$776 &  .153$\pm$.048 &  284$\pm$137 &  .005$\pm$.0 &  {\bf .004$\pm$.0} & 63.4$\pm$109.3 &  .005$\pm$.0 & .207$\pm$.043 & .009$\pm$.001 &  {\bf .004$\pm$.0} &  .005$\pm$.001 &  .005$\pm$.001\\
    ISOLET       		&.304$\pm$.047 &  .104$\pm$.007 &  .144$\pm$.005 &  .016$\pm$.0 &  .019$\pm$.001 &  .154$\pm$.032 &  .015$\pm$.0 & .099$\pm$.015 & .035$\pm$.005 &  {\bf .013$\pm$.0} &  .015$\pm$.0 &  .017$\pm$.0\\
    MNIST		 		& .305$\pm$.0 &  .067$\pm$.001 &  .134$\pm$.004  & .037$\pm$.002 &  .029$\pm$.002   &  .128$\pm$.003 &  .028$\pm$.001 & .101$\pm$.003 &  .055$\pm$.005 &  {\bf .019$\pm$.0} & {\bf .019$\pm$.0} &  .025$\pm$.001\\
    MNIST-Fashion 		& 11.4$\pm$22.5 & .109$\pm$.007 & .139$\pm$.010   &  .023$\pm$.0 &  .027$\pm$.003  &  .458$\pm$.657 &  .022$\pm$.0 &  .105$\pm$.006       & .025$\pm$.001 &  {\bf .019$\pm$.0} &  {\bf .019$\pm$.0} &  .022$\pm$.0\\
	USPS         		& 2.99$\pm$.73 &  1.28$\pm$.16 & 1.07$\pm$.08 & .027$\pm$.003 &  .032$\pm$.002 &  1.07$\pm$.07 &  .018$\pm$.002 & 4.05 $\pm$.63&  .017$\pm$.003 & .012$\pm$.001 &  {\bf .011$\pm$.001} &  .021$\pm$.001\\
    GLIOMA 				&.226$\pm$.033 &  .259$\pm$.030 &  .347$\pm$.044 &  {\bf .065$\pm$.008} & .072$\pm$.004 &  .249$\pm$.019 &  .067$\pm$.011 & .211$\pm$.065 &  .068$\pm$.012 &  .066$\pm$.011 & .069$\pm$.010 & .141$\pm$.032\\
    leukemia     		&10.7$\pm$5.7 &  8.50$\pm$1.41 &  12.7$\pm$4.0 &  7.78$\pm$2.42 & $\backslash$ & 14.1$\pm$4.0 &  6.30$\pm$1.22 & 12.3$\pm$2.7 & {\bf 6.14$\pm$1.07} & 398$\pm$736 &  7.01$\pm$.94 &  9.07$\pm$3.56\\
	pixraw10P        	& 31.1$\pm$29.8 &  .554$\pm$.155 &  .645$\pm$.400 &  .006$\pm$.002 &  $\backslash$ &  .163$\pm$.041 &  .003$\pm$.001 & 1.357$\pm$.700 &  .009$\pm$.004 &  .013$\pm$.011 &  .005$\pm$.004 &  {\bf .002$\pm$.001}\\
	ProstateGE        	&1.36$\pm$.42 &  .404$\pm$.310 & 3.91$\pm$2.05 &  .242$\pm$.094 &  $\backslash$ & 3.00$\pm$3.36 &  .142$\pm$.039 & .273 $\pm$.093 &  .146$\pm$.026 &  .202$\pm$.137 &  .144$\pm$.039 & {\bf .068$\pm$.016}\\
    warpAR10P    		&1.28$\pm$.61 &  4.73$\pm$3.95 &  .597$\pm$.118 &  .039$\pm$.007 &  .086$\pm$.029 &  1.08$\pm$.33 &  .036$\pm$.005 & 3.68$\pm$1.15 & .045$\pm$.006 &  .074$\pm$.027 &  .040$\pm$.005 &  {\bf .033$\pm$.005}\\
	SMK$\_$CAN$\_$187   &5.87$\pm$1.08 &  .127$\pm$.020 & $3.52\pm.62$ &  .114$\pm$.022 & $\backslash$ &  6.24$\pm$1.00 &  .110$\pm$.016 & 6.84$\pm$2.53 &  .102$\pm$.012 &  .100$\pm$.015 &  .105$\pm$.019 & {\bf .097$\pm$.021}\\
   	arcene       		&.410$\pm$.250 &  .045$\pm$.001 &  1493$\pm$267  &  .055$\pm$.043 & $\backslash$ &  1.86$\pm$.62 &  .035$\pm$.009 &478$\pm$205& .030$\pm$.012 &  .027$\pm$.001 &  .025$\pm$.001 &  {\bf .023$\pm$.001}\\
    \bottomrule
  \end{tabular}}}
  \caption{Linear reconstruction error with selected features by different algorithms. The \lq\lq $\backslash$\rq\rq\, mark denotes the case with prohibitive running time, where the algorithm ran for more than one week without getting the result and thus was stopped. }
\label{table2}
\end{table*}

\begin{table*}[!htbp] 
\centering
\resizebox{1.90\columnwidth}{!}{
  {\setlength{\aboverulesep}{0pt}
   \setlength{\belowrulesep}{0pt}
  \begin{tabular}{l!{\vrule width0.8pt}cccccccccc|cc}
    \toprule
    \multirow{2}*{{\Large Dataset}}&\multirow{2}*{{\Large LS}}&\multirow{2}*{{\Large SPEC}}&\multirow{2}*{{\Large NDFS}}&\multirow{2}*{{\Large AEFS}}&\multirow{2}*{{\Large UDFS}}&\multirow{2}*{{\Large MCFS}}&\multirow{2}*{{\Large PFA}}&\multirow{2}*{{\Large Inf-FS}}&\multirow{2}*{{\Large AgnoS-S}}&\multirow{2}*{{\Large CAE}}&\multicolumn{2}{c}{{\Large FAE}}\\
    					& 	  & 	& 								& 			& 	  &	  	& 			& &			& 			&Opt1		&Opt2\\
    \midrule
    Mice Protein 		&17.3$\pm$4.0 &  13.7$\pm$3.6 &  15.6$\pm$10.5 &  88.5$\pm$4.0 &  {\bf 95.1$\pm$4.0} &  17.4$\pm$5.9 &  92.8$\pm$2.9 & 19.3$\pm$6.9 & 63.2$\pm$37.2 &  66.9$\pm$4.8 &  87.8$\pm$7.3 &  78.6$\pm$18.2\\
    COIL-20      		& 16.0$\pm$3.4 &  16.4$\pm$2.6 &  16.8$\pm$3.9 &  99.3$\pm$.2 &  97.6$\pm$2.3 &  10.5$\pm$2.6 &  99.4$\pm$.4 & 34.4$\pm$9.0 &  83.5$\pm$14.5 &  98.8$\pm$.5 &   {\bf 99.6$\pm$.3} &  99.0$\pm$1.0\\
    Activity     		& 29.0$\pm$1.4 &  20.3$\pm$1.1 &  18.2$\pm$1.2 &  88.7$\pm$1.7 &  91.7$\pm$1.5 &  21.6$\pm$4.6 &  88.8$\pm$1.4 & 24.1$\pm$5.0 & 74.9$\pm$10.7 &  {\bf 91.9$\pm$1.0} &  91.4$\pm$1.1 &  88.8$\pm$1.3\\
    ISOLET       		&11.1$\pm$.7 &  3.4$\pm$.9 &  8.4$\pm$1.6 &  82.9$\pm$2.1 &  73.9$\pm$5.8 &  6.0$\pm$1.2 &  86.5$\pm$1.3 & 14.0$\pm$1.9 &  36.2$\pm$17.0 &  87.9$\pm$.6 &  {\bf 89.0$\pm$.6} &  87.0$\pm$1.3\\
    MNIST		 		& 12.4$\pm$1.0 &  11.2$\pm$1.0 &  10.5$\pm$1.9 & 80.2$\pm$2.6 & 88.1$\pm$1.6 &  13.0$\pm$4.1 &  88.5$\pm$2.0 & 13.2$\pm$1.3 &  43.5$\pm$15.6 &  92.5$\pm$.4 &  {\bf 92.9$\pm$.7} &  90.8$\pm$.9	\\
    MNIST-Fashion		& 17.1$\pm$2.8 &  27.2$\pm$1.7 &  15.2$\pm$4.5  &  79.4$\pm$1.5 &   79.3$\pm$1.2 &  16.3$\pm$.9 &  80.3$\pm$10.5 &    18.8$\pm$3.8    &  78.4$\pm$1.3  &  82.3$\pm$1.0 &  {\bf 82.5$\pm$.6} &  80.8$\pm$.9\\
    USPS         		& 34.2$\pm$4.8 &  36.2$\pm$6.9 & 10.6$\pm$1.8 &  94.9$\pm$.7 &  94.5$\pm$.3 &  12.3$\pm$3.4 &  95.7$\pm$.3 & 18.2$\pm$4.2 & 95.6$\pm$.4 &  96.2$\pm$.4 & {\bf 96.3$\pm$.2} &  96.2$\pm$.3\\
    GLIOMA 				&46.0$\pm$17.4 &  22.0$\pm$16.0 &  34.0$\pm$8.0 &  68.0$\pm$13.3 &  76.0$\pm$20.6 &  46.0$\pm$8.0 &  66.0$\pm$8.0 & 42.0$\pm$11.7 &  62.0$\pm$7.5 &  70.0$\pm$16.7 &  {\bf 76.0$\pm$8.0} &  72.0$\pm$16.0\\
    leukemia     		&52.0$\pm$12.2 &  58.7$\pm$8.8 &  57.3$\pm$13.7 &  76.0$\pm$12.4 & $\backslash$ & 54.7$\pm$10.7 &  72.0$\pm$14.9 & 56.0$\pm$6.8 &  72.0$\pm$14.9 &  65.3$\pm$12.9 &  80.0$\pm$11.2 &  {\bf 80.0$\pm$6.0}\\
    pixraw10P        	&51.0$\pm$8.6 &  9.0$\pm$5.8 &  19.0$\pm$8.0 &  {\bf 100.0$\pm$0.0} &  $\backslash$ &  11.0$\pm$5.8 &  {\bf 100.0$\pm$0.0} &  41.0$\pm$23.5 & {\bf 100.0$\pm$0.0} &  99.0$\pm$2.0 &  {\bf 100.0$\pm$0.0} &  {\bf 100.0$\pm$0.0}\\
    ProstateGE        	&43.8$\pm$9.2 &  55.2$\pm$13.7 & 56.2$\pm$17.4 & 85.7$\pm$8.5 &  $\backslash$ & 46.7$\pm$11.0 &  81.0$\pm$10.0 & 57.1 $\pm$15.7 &  81.9$\pm$4.7 &  78.1$\pm$9.8 &  82.9$\pm$5.7 & {\bf 85.7$\pm$4.2}\\
    warpAR10P    		&11.5$\pm$5.4 &  7.7$\pm$4.9 &  12.3$\pm$7.1 &  76.9$\pm$5.4 &  64.6$\pm$8.9 &  15.4$\pm$5.4 &  {\bf 82.3$\pm$6.7} & 10.8$\pm$3.8 & 71.5$\pm$7.1 &  70.0$\pm$10.7 &  71.5$\pm$3.1 &  63.8$\pm$7.9\\
    SMK$\_$CAN$\_$187  	&55.3$\pm$7.1 &  65.8$\pm$6.9 & $53.2\pm14.6$ &  61.1$\pm$4.5 & $\backslash$ &  50.5$\pm$4.2 &  70.0$\pm$2.7 & 47.4$\pm$2.9 &  71.6$\pm$5.6 &  69.5$\pm$8.3 &  {\bf 72.1$\pm$6.6} & 71.1$\pm$8.5\\
    arcene       		&59.5$\pm$7.0 &  51.5$\pm$9.8 &  57.5$\pm$7.3 &  81.0$\pm$5.2 & $\backslash$ &  56.5$\pm$4.6 &  80.0$\pm$5.7 & 59.5$\pm$5.3 & 80.0$\pm$4.5 &  74.5$\pm$8.9 &  {\bf 84.0$\pm$6.6} &  80.0$\pm$4.5\\
    \bottomrule
  \end{tabular}}}
  \caption{Classification accuracy (\%) with selected features by different algorithms. The mark \lq\lq $\backslash$ \rq\rq\, is used similarly to Table~\ref{table2}.}
\label{table3}
\end{table*}

Further, we compare the behaviors of FAE with respect to $k$ with those of the baseline algorithms. By varying $k$ on ISOLET and arcene, we obtain the corresponding linear reconstruction errors and classification accuracies. We plot the results in Figure~\ref{fig:0712}{\footnote{For better visualization, we ignore the algorithms with large linear reconstruction errors.}}. The results show that FAE performs better and more stable than other algorithms in most cases. 

\begin{figure}[!htbp]
\centering
\subfigure[ISOLET]{
\centering
{
\begin{minipage}[t]{0.99\linewidth}
\centering
\centerline{\includegraphics[width=1.0\textwidth]{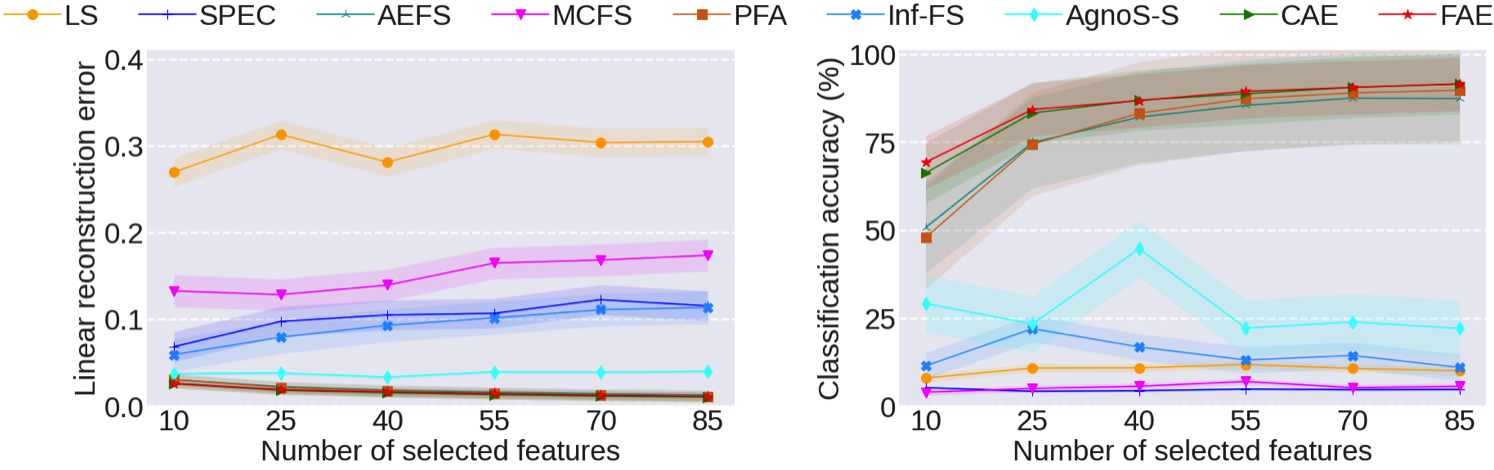}}
\end{minipage}%
}%
}%
\hspace{.0in}
\subfigure[arcene]{
\centering
{
\begin{minipage}[t]{0.99\linewidth}
\centering
\centerline{\includegraphics[width=1.0\textwidth]{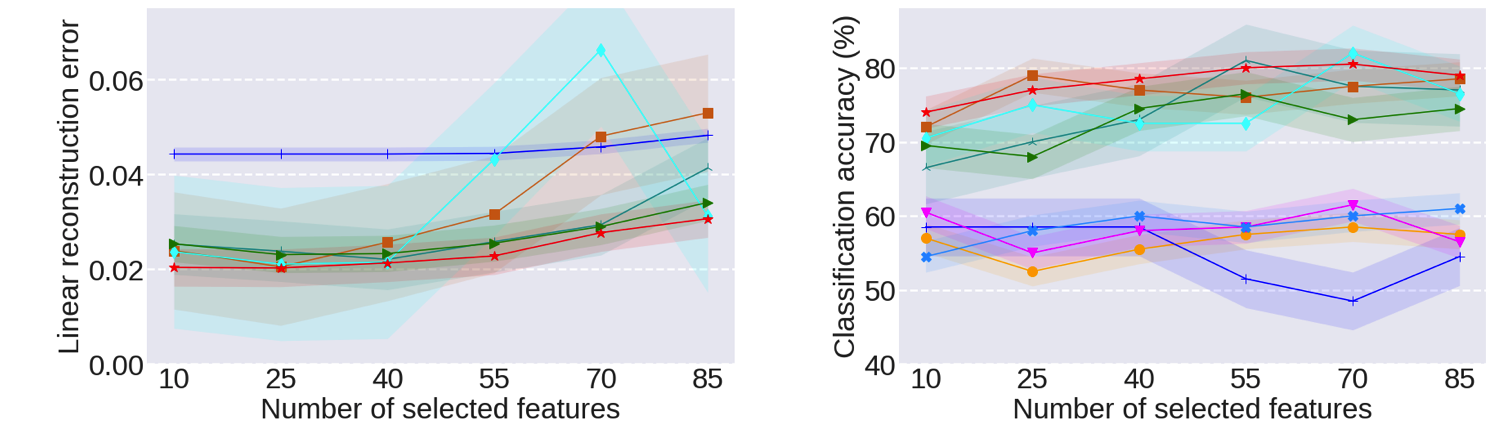}}
\end{minipage}%
}%
}%
\centering
\caption{Reconstruction and classification results versus $k$.}
\label{fig:0712}
\end{figure} 

\subsubsection{Feature Importance}
To examine the importance of the features selected by FAE, we rank and partition them into two equal groups, that is, each group has $25$ features. The results are shown in Figure~\ref{fig:featurerank}. We can observe that, the classification accuracy of the first group is generally better than the second group. However, since FAE is unsupervised, some selected features that are essential for reconstruction might not be important for classification. 

\begin{figure}[!htbp]
\centering
{
\centerline{\includegraphics[width=7cm,height=2.4cm]{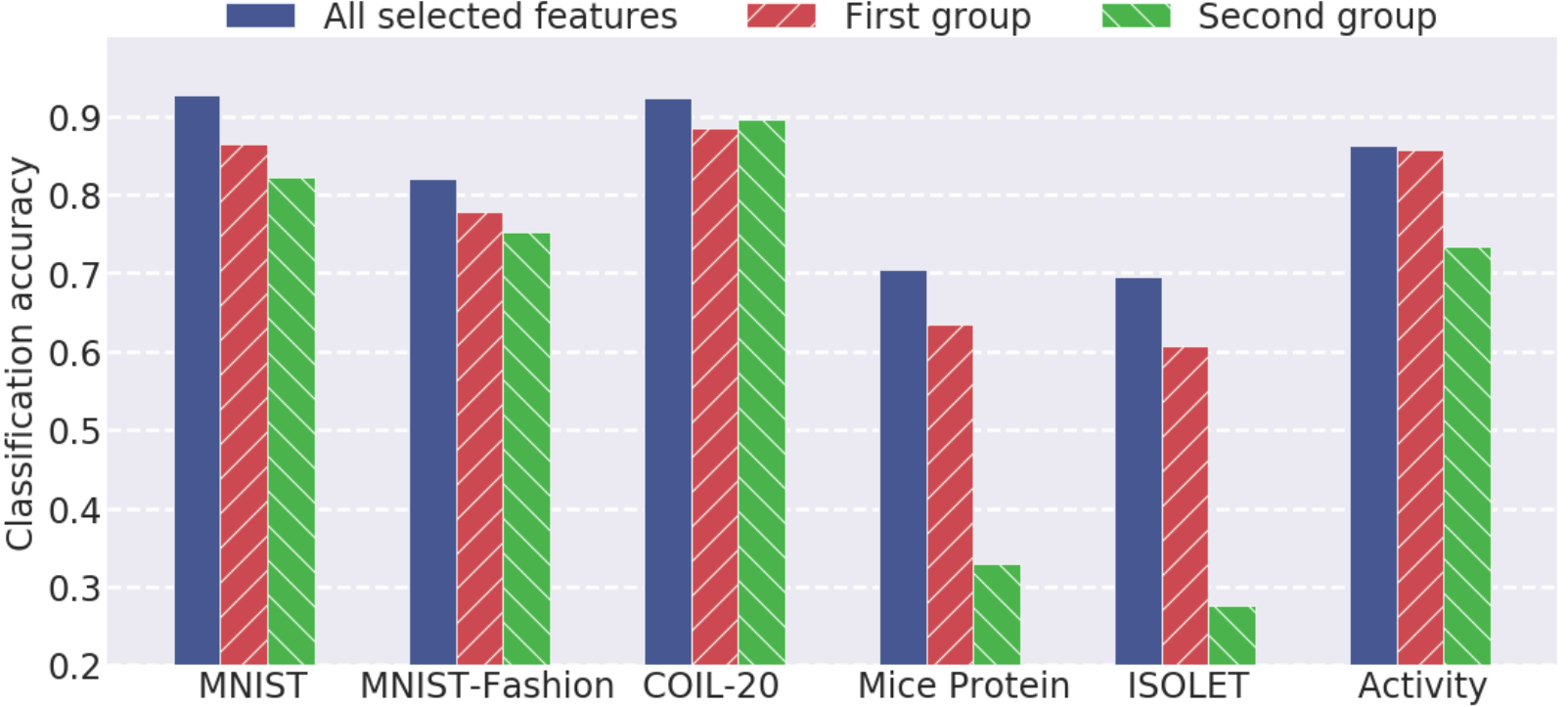}}
}%
\centering
\caption{{\footnotesize{Analyses of feature importance with $k=50$.}}}
\label{fig:featurerank}
\end{figure}
\subsection{Computational Complexity}
Experimentally, the computational time of our algorithm (\ref{FAE}) is about twice that of sparse AE. FAE has only an additional sub-NN compared to sparse AE and shares parameters with the global-NN. Also, the fitting error term of sub-NN is quadratic and similar to sparse AE's fitting error term. Thus, the overall computational complexity of FAE is of the same order as sparse AE. 

\subsection{Analysis of L1000 Gene Expression}
It is expensive to measure all gene expressions. To reduce the cost, researchers from the LINCS program{\footnote{See http://www.lincsproject.org/LINCS/}} have found that a carefully selected set of genes can capture most gene expression information of the entire human transcriptome because the expression of genes is usually correlated under different conditions. Based on the selected genes, a linear regression model was used to infer the gene expression values of the remaining genes \citep{Chen}. Recently, \citet{Abubakar} have used CAE to further reduce the number of genes to 750 to achieve a similar linear reconstruction error about $0.3$ to the original $943$ landmark genes of L1000.

\begin{figure}[!htbp]
\centering
\subfigure[]{
\centering
{
\begin{minipage}[t]{0.46\linewidth}
\centering
\centerline{\includegraphics[width=1.02\textwidth]{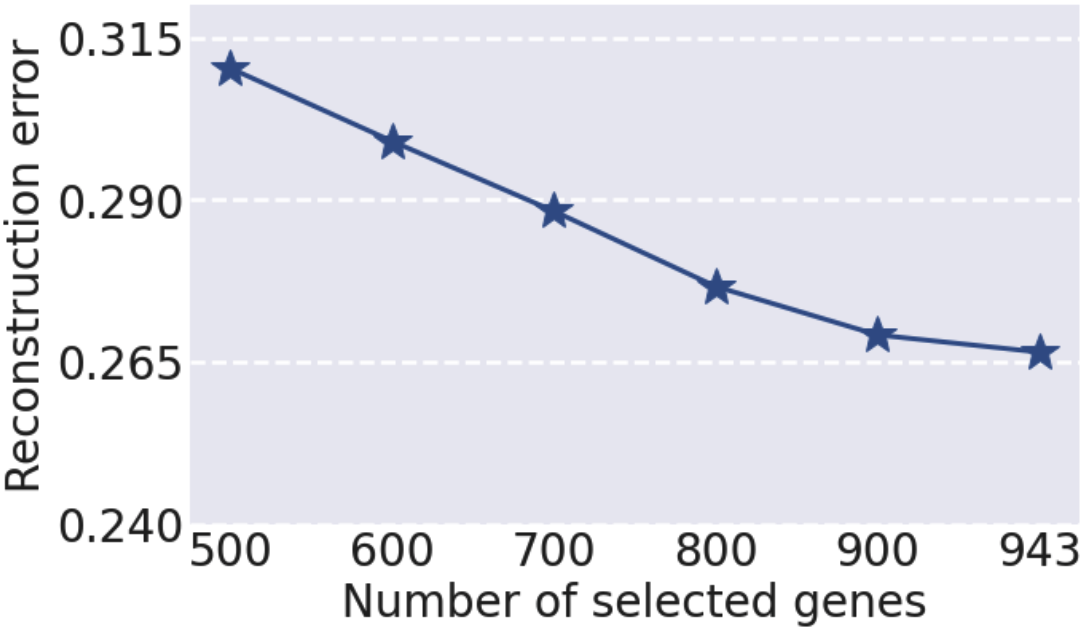}}
\end{minipage}%
}%
}%
\hspace{-0.06in}
\subfigure[]{
\centering
{
\begin{minipage}[t]{0.46\linewidth}
\centering
\centerline{\includegraphics[width=1.02\textwidth]{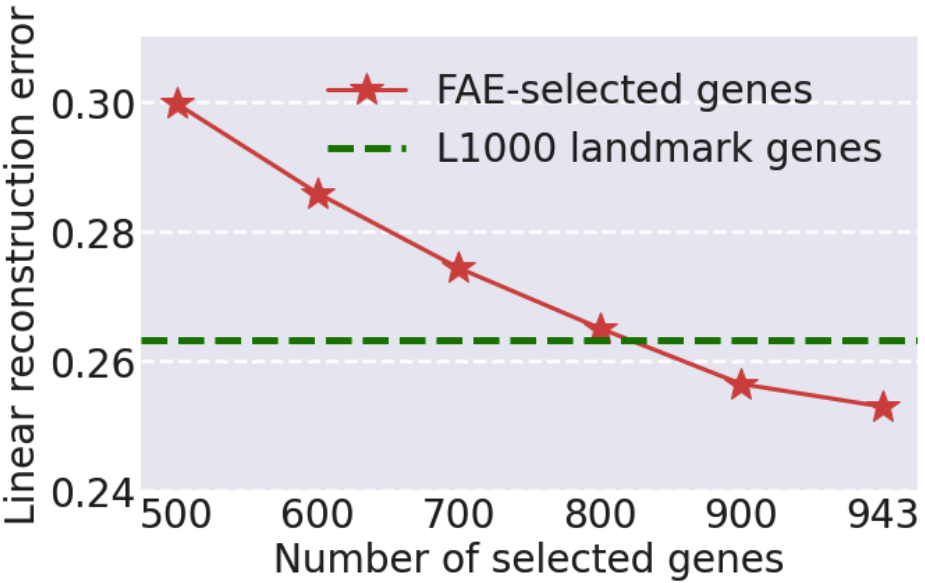}}
\end{minipage}%
}%
}%
\centering
\caption{Gene selection by using FAE for GEO. (a) Reconstruction error by FAE; (b) Reconstruction error by using the linear regression model on L1000 landmark genes and FAE-selected genes.}
\label{fig:genes}
\end{figure} 

Now we apply FAE on the preprocessed GEO to select varying numbers of representative genes from $500$ to $943$. Figure~\ref{fig:genes} (a) shows that, by using $600$ genes, FAE achieves a reconstruction error better than that with $750$ selected genes by CAE. However, CAE uses a slightly different number of samples with ours. For consistency, we mainly compare FAE with L1000. We compute the reconstruction error by using the linear regression model on the genes selected by FAE and the landmark genes of L1000, and the results in MSE are depicted in Figure~\ref{fig:genes} (b). Evidently, using $800$ genes by FAE achieves a similar reconstruction to L1000. Thus, FAE reduces the number of genes by about $15$\% compared to L1000. The selected genes by FAE are displayed in Figure~\ref{FAEselectedgene}. It is observed that, with different numbers of selected genes, a few genes sometimes are selected and sometimes not, which may be attributed to the significant correlations among genes. In addition, when selecting the same number of 943 genes, only $90$ genes selected by CAE are among the landmark genes, while $121$ by FAE are among the landmark genes. These results indicate that L1000 landmark genes can be significantly enhanced in representation power. 
\begin{figure}[!htbp]
\centering
{
\centerline{\includegraphics[width=0.43\textwidth]{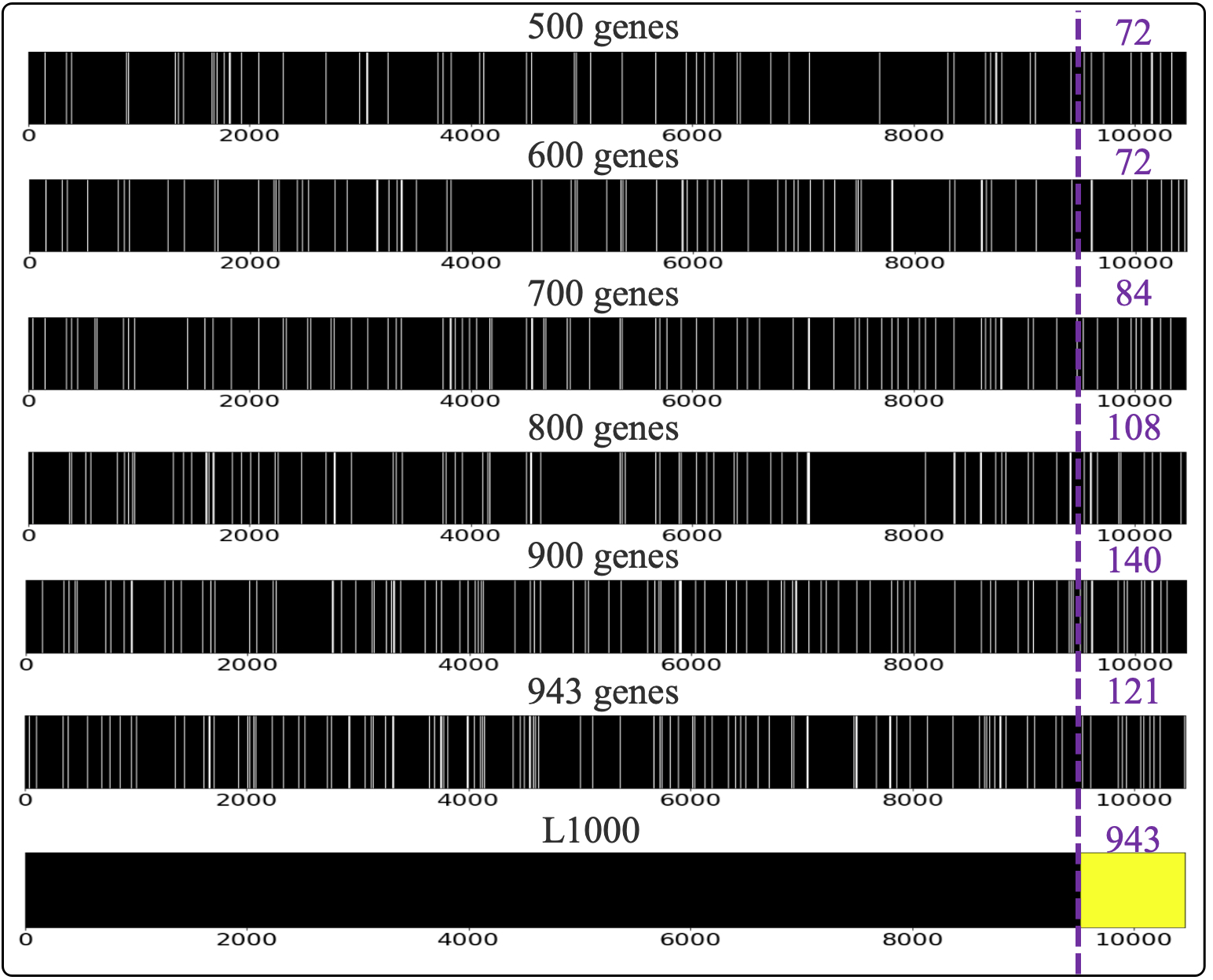}}
}%
\centering
\caption{Comparison of different numbers of selected genes with 943 landmark gene. The white lines denote those selected genes by FAE. The purple dashed line separates $943$ landmark genes (color coded in yellow) from the other genes. The purple numbers $72$, $72$, $84$, $108$, $140$, and $121$ denote respectively the numbers of overlapping genes between the landmark genes and those selected by FAE with $k$ being 500, 600, 700, 800, 900, and 943.}
\label{FAEselectedgene}
\end{figure} 

\section{An Application of FAE}
FAE is applicable and easily extensible to different tasks. Here we show an application of exploiting multiple hierarchical subsets of the key features.

\subsection{$h$-HFAE}
For an image, usually there are many pixels highly correlated with each other. Thus, the subsets of key features might not be unique; indeed, there often exists more than one subset of informative features that can recover the original data well. Excavating these potential subsets of meaningful features is conductive to facilitate data compression~\citep{Sousa} and better understand the structure and inter-relationship of the features. Yet, almost all existing feature selection approaches have little ability to explore these potential subsets. To achieve such an ability, we develop an application in the framework of FAE, which selects multiple non-overlapping subsets of representative features. For clarity, we formalize it as follows:
\begin{equation}{\label{HFAE}}
\begin{array}{l}
\displaystyle\min_{\mathbf{W}_{\mathrm{I}}^{\mathrm{max}_{k,i}},\mathbf{W}_{\mathrm{E}},\mathbf{W}_{\mathrm{D}}}\|\mathbf{X}-((\mathbf{X}\mathbf{W}_{\mathrm{I}})\mathbf{W}_{\mathrm{E}})\mathbf{W}_{\mathrm{D}}\|_{\mathrm{F}}^2+\lambda_{0}\|\mathbf{W}_{\mathrm{I}}\|_{1}\\
+\sum_{i=1}^{h}\lambda_i\|\mathbf{X}-((\mathbf{X}\mathbf{W}_{\mathrm{I}}^{\mathrm{max}_{k,i}})\mathbf{W}_{\mathrm{E}})\mathbf{W}_{\mathrm{D}}\|_{\mathrm{F}}^2,\,\, \mathrm{s.t.}\,\,\mathbf{W}_{\mathrm{I}}\geqslant 0,
\end{array}
\end{equation}
where $\mathbf{W}_{\mathrm{I}}^{\mathrm{max}_{k,1}}=$ Diag$(\mathbf{w}^{\mathrm{max}_{k,1}})$, $\mathbf{W}_{\mathrm{I}}^{\mathrm{max}_{k,i}}=$ Diag$((\mathbf{w}/\mathbf{w}^{\mathrm{max}_{k,{i-1}}})^{\mathrm{max}_{k,i}})$, $i=2,\ldots,h$, $h$ is the number of desired subsets of relevant features, $\lambda_i, i=0,\ldots,h$, are hyper-parameters, and $(\mathbf{w}/\mathbf{w}^{\mathrm{max}_{k,{i-1}}})^{\mathrm{max}_{k,i}}$ is an operation to retain the $i$-th group of $k$ largest entries from $\mathbf{w}$ while making zero all the other entries including the $(i-1)$ groups of $k$ largest entries of $\mathbf{w}^{\mathrm{max}_{k,1}},\mathbf{w}^{{\mathrm{max}_{k,2}}}, \ldots$, and $\mathbf{w}^{{\mathrm{max}_{k,{i-1}}}}$. In (\ref{HFAE}), the first two terms estimate the importance of each input feature globally; then, the remaining terms organize the top $kh$ features into $h$ hierarchical subsets in descending order of importance values, with each subset having $k$ features. These $h$ subsets are selected by using $h$ sub-NNs, which work together in an orchestrated way: The $(i+1)$-th sub-NN exploits the $(i+1)$-th hierarchical subset of features by leaving out the $i$ subsets of features found by the previous $i$ sub-NN(s). For notational convenience, we denote this application for identifying multiple hierarchical subsets of features by $h$-HFAE. 

To verify the effectiveness of $h$-HFAE, we set $h=3$ and apply it to MNIST. The reconstruction and classification results together with those of FAE are shown in Figure~\ref{fig:05}, where we set the hyper-parameters $\lambda_0$, $\lambda_1$, $\lambda_2$, and $\lambda_3$ in (\ref{HFAE}) to be $0.05$, $1.5$, $2$, and $3$, respectively. With 50 selected features per group, different hierarchies of $3$-HFAE achieve almost the same accuracy; with 36 selected features per group, the third group of features from $3$-HFAE-$\mathrm{H}_3^3$ has slightly worse accuracy than the other two groups. This result implies that 50 selected features per group are more stable for $3$-HFAE. For reconstruction error, the vanilla version of FAE is the best among all results.

\begin{figure}[!htbp] 
  \centering
	\includegraphics[width=0.465\textwidth]{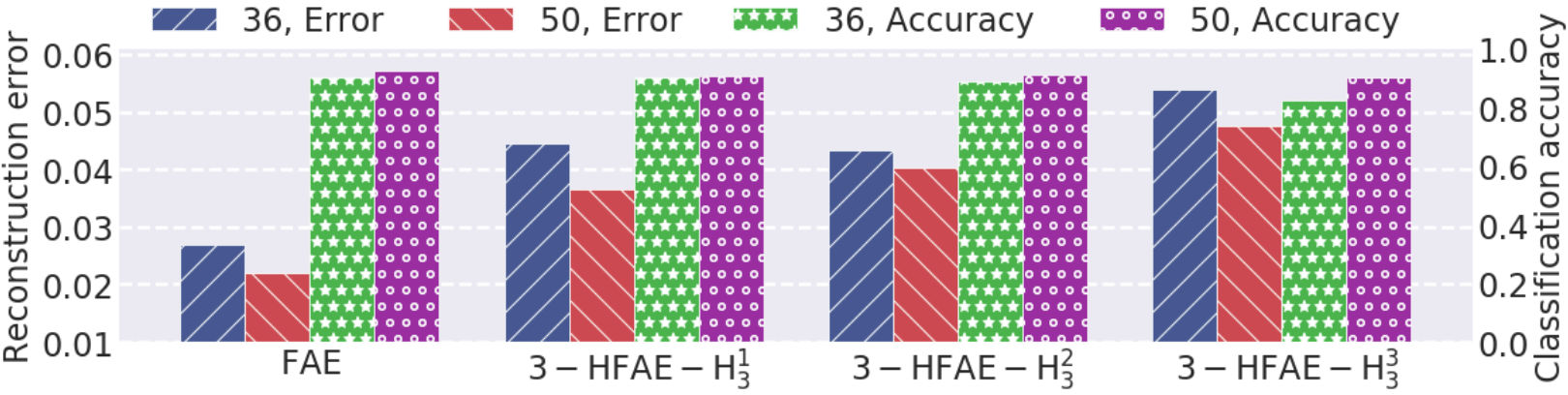}
  \caption{Reconstruction and classification results of FAE and $3$-HFAE on MNIST. $3$-HFAE-$\mathrm{H}_3^1$, $3$-HFAE-$\mathrm{H}_3^2$, and $3$-HFAE-$\mathrm{H}_3^3$ denote respectively the first three hierarchical subsets of selected features.}
\label{fig:05}
\end{figure}
CAE~\citep{Abubakar} displays the relationships of the top $3$ selected features at each node of the concrete selector layer; however, the second and third top features might be insignificant due to the potentially trivial average probability ($\leqslant0.01$). Different from CAE, $h$-HFAE uses the weights to assign the features into different hierarchical subsets for selection and exploration. In the Supplementary Material, we demonstrate that for $3$-HFAE there exists a considerable degree of similarity between different hierarchical subsets of selected features. Thus, $h$-HFAE can reveal the redundancy or high correlations among features.

\section{Conclusions}
In this paper, we propose a new framework for unsupervised feature selection, which extends AE by adding a simple one-to-one layer and a sub-NN to achieve both global exploring of representative abilities of the features and local mining for their diversity. Extensive assessment of the new framework has been performed on real datasets. Experimental results demonstrate its superior performance over contemporary methods. Moreover, this new framework is applicable and easily extensible to other tasks and we will further extend it in our future work.

\section{Acknowledgments}
This work is supported in part by NSF OIA2040665, NIH R56NS117587, and R01HD101508. We sincerely thank the anonymous reviewers for their valuable comments.

\bibliography{references}

\end{document}